\documentclass{article}

\usepackage{microtype}
\usepackage{graphicx}
\usepackage{subfigure}
\usepackage{svg}
\svgpath{{./figures/}}
\usepackage{booktabs} 
\usepackage{algorithm}
\usepackage{multirow}

\usepackage{amsmath}
\usepackage{amssymb}

\usepackage{hyperref}

\usepackage[accepted]{arxiv}

\usepackage{amsmath}
\usepackage{amssymb}
\usepackage{mathtools}
\usepackage{amsthm}
\usepackage{bm}

\usepackage[capitalize,noabbrev]{cleveref}

\theoremstyle{plain}

\theoremstyle{definition}

\theoremstyle{remark}

\usepackage[textsize=tiny]{todonotes}

\icmltitlerunning{ Enhancing Auto-regressive Chain-of-Thought through Loop-Aligned Reasoning}

\begin{document}

\twocolumn[
\icmltitle{ Enhancing Auto-regressive Chain-of-Thought through Loop-Aligned Reasoning}



\icmlsetsymbol{equal}{*}

\begin{icmlauthorlist}
\icmlauthor{Qifan Yu}{pku}
\icmlauthor{Zhenyu He}{pku}
\icmlauthor{Sijie Li}{pku}
\icmlauthor{Xun Zhou}{byte}
\icmlauthor{Jun Zhang}{byte}
\icmlauthor{Jingjing Xu}{byte}
\icmlauthor{Di He}{pku}
\end{icmlauthorlist}

\icmlaffiliation{pku}{Peking University}
\icmlaffiliation{byte}{ByteDance Inc.}


\icmlkeywords{Machine Learning, ICML}

\vskip 0.3in
]



\printAffiliationsAndNotice{}  

\begin{abstract}
Chain-of-Thought (CoT) prompting has emerged as a powerful technique for enhancing language model's reasoning capabilities. However, generating long and correct CoT trajectories is challenging. 
Recent studies have demonstrated that Looped Transformers possess remarkable length generalization capabilities, but their limited generality and adaptability prevent them from serving as an alternative to auto-regressive solutions. To better leverage the strengths of Looped Transformers, we propose \textbf{RELAY} (\underline{\textbf{RE}}asoning through \underline{\textbf{L}}oop \underline{\textbf{A}}lignment iterativel\underline{\textbf{Y}}).
Specifically, we align the steps of Chain-of-Thought (CoT) reasoning with loop iterations and apply intermediate supervision during the training of Looped Transformers. This additional iteration-wise supervision not only preserves the Looped Transformer's ability for length generalization but also enables it to predict CoT reasoning steps for unseen data. Therefore, we leverage this Looped Transformer to generate accurate reasoning chains for complex problems that exceed the training length, which will then be used to fine-tune an auto-regressive model. We conduct extensive experiments, and the results demonstrate the effectiveness of our approach, with significant improvements in the performance of the auto-regressive model. Code will be released at \url{https://github.com/qifanyu/RELAY}.
\end{abstract}

\section{Introduction}
Reasoning plays a central role in shaping effective decision-making processes and guiding problem-solving strategies in artificial intelligence systems. For large language models (LLMs), the most effective way to achieve reasoning is through Chain-of-Thought~\cite{wei2022chain,khot2022decomposed}, which generates all intermediate steps token by token until the final answer is reached. However, generating the correct reasoning process using LLMs is challenging. On one hand, the Chain-of-Thought process can be very long, sometimes growing polynomially with respect to the prompt length~\cite{feng2024towards,merrill2024the}. When the reasoning length exceeds the training data length, it encounters the length generalization problem, where accuracy can drop significantly~\cite{xiao2023conditions,jin-etal-2024-impact}. On the other hand, web data is often noisy, and learning from incorrect trajectories can lead to incorrect answers. While synthetic data could mitigate this issue~\cite{lightman2024lets}, it requires significant human effort and knowledge to generate and curate. 

\begin{figure}[t]
    \centering
    \includegraphics[width=\linewidth]{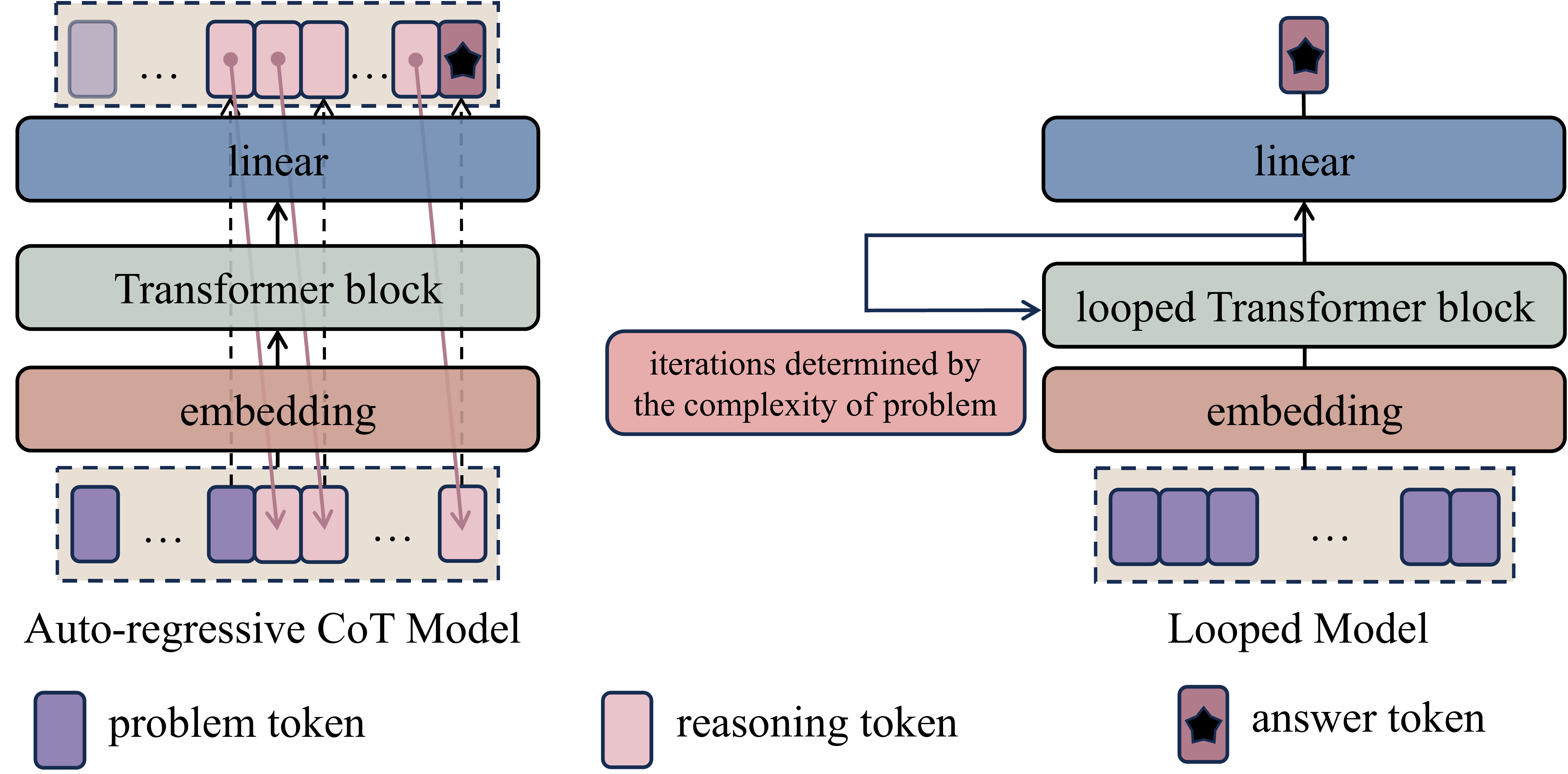}
    \caption{Visualization of Chain-of-Thought (CoT) and looping process. As the complexity of problem increases, in the auto-regressive CoT model, the number of reasoning tokens escalates. In contrast, in the looped model, the number of iterations of the loop block increases. } 
    \label{fig:loop}
    \vspace{-14pt}
\end{figure}

Recently, an alternative framework has gained attention, known as the looped Transformer~\cite{giannou2023looped}. In general, the looped Transformer is a standard Transformer model with cross-block parameter sharing, like AlBERT~\cite{lan2020albert}. In this framework, the input prompt (i.e., the problem) is processed through repeated iterations of the same block, with the number of iterations adaptively determined by the problem complexity. See Figure~\ref{fig:loop} for an illustration. Several preliminary results~\cite{fan2024looped} show that the looped Transformer model has better length generalization capabilities, partially because the increase in problem complexity (e.g., problem length) is not as significant as in the Chain-of-Thought steps. 


However, the success of this approach comes with some practical limitations. While determining appropriate loop iterations is feasible for reasoning tasks, it becomes problematic in general tasks, such as translation and summarization. Furthermore, although the looped Transformer can handle specific reasoning tasks, it remains unclear whether it possesses the capability to manage multiple reasoning tasks within a single model. Given these concerns, a natural question arises: if the looped Transformer is a general reasoner, can we explore ways to integrate its capabilities into the Chain-of-Thought framework of standard auto-regressive models? This integration would allow us to leverage the looped Transformer's strong performance on complex reasoning problems while preserving the versatility that allows auto-regressive models to excel in diverse language tasks.

In this paper, we introduce \textbf{RELAY} (\underline{\textbf{RE}}asoning through \underline{\textbf{L}}oop \underline{\textbf{A}}lignment iterativel\underline{\textbf{Y}}), a novel framework that leverages looped Transformer's superior capabilities to help auto-regressive models handle longer reasoning chains. At its core, our approach centers on two key innovations. First, we demonstrate empirically that a single looped Transformer model can serve as a general reasoner across multiple tasks while maintaining strong length generalization abilities. Second, we propose an iteration-wise alignment between the looped Transformer and Chain-of-Thought reasoning steps, enabling the looped model to generate accurate reasoning chains for problems beyond training length. These generated reasoning chains can then serve as training data for auto-regressive models, establishing a bridge between the two architectural paradigms.

We conduct extensive experiments demonstrating that our approach significantly improves the reasoning abilities of auto-regressive Transformers through high-quality generated reasoning chains.

\section{Related Work}

\subsection{Auto-regressive LLM with Chain-of-Thought}

Chain-of-Thought (CoT) has emerged as a powerful technique for enhancing language models' reasoning capabilities both empirically~\cite{wei2022chain,khot2022decomposed} and theoretically~\cite{feng2024towards,merrill2024the}, especially in latest models such as OpenAI O1\footnote{https://openai.com/o1}, DeepSeek r1~\cite{deepseekai2025deepseekr1} and Qwen QwQ\footnote{https://qwenlm.github.io/blog/qwq-32b-preview}. By generating intermediate reasoning steps token by token, these models effectively decompose complex problems into sequential subprocesses. However, two critical challenges persist. First, obtaining high-quality CoT training data remains time-consuming and labor-intensive~\cite{lightman2024lets}, especially for problems requiring sophisticated reasoning chains. Second, the generation and understanding of extended reasoning sequences can be problematic~\cite{xiao2023conditions,jin-etal-2024-impact,mao2024lift}.

\subsection{Looped Transformer}
Research on looped Transformers has evolved significantly over recent years. The initial studies by \citet{dehghani2018universal} and \citet{lan2020albert} demonstrated the effectiveness of parameter sharing across layers in supervised learning and BERT pretraining. This line of research has since expanded in both theoretical and practical directions. On the theoretical front, \citet{giannou2023looped} and \citet{xu2024expressive} established fundamental properties of looped Transformers, proving their Turing completeness and characterizing their approximation capabilities. \citet{gatmiry2024can} further advanced this understanding by showing how to incorporate inductive biases for learning iterative algorithms, particularly in the context of multi-step gradient descent for in-context learning. Empirically, looped Transformers have shown promising results across various applications. \citet{yang2024looped} demonstrated their parameter efficiency in data-fitting tasks, while \citet{luca2024simulation} and \citet{chen2024bypassing} revealed their potential in graph algorithm simulation and in-context learning enhancement. Notably, \citet{fan2024looped} established their superior length generalization capabilities in RASP-L tasks. In the domain of algorithm learning, \citet{gao2024expressive} introduced AlgoFormer, a framework that leverages looped Transformers for algorithm representation and learning. While these works have extensively explored various aspects of looped Transformers, our work takes a distinct direction. We specifically focus on leveraging the better length generalization of looped Transformers for helping standard auto-regressive Transformers.

\subsection{Approaches for Length Generalization}

The capability of Transformers to generalize to longer sequence is influenced by their positional encodings~\citep{alibi}. Recent research has pursued two primary directions to enhance length generalization capabilities of LLMs. The first focuses on developing advanced relative positional encoding schemes~\citep{raffel2020exploring,alibi,chi2022kerple,sun2022length,chi2023dissecting,li2024functional}, while the second explores modifications to positional representations through index granularity adjustments~\cite{chen2023extending,peng2024yarn} and strategic index shifting~\cite{ruoss2023randomized,zhu2024pose}. These works are orthogonal to the central contributions of this paper. 
A parallel line of work focuses on improving the reasoning capabilities of LLMs through better training data. These methods typically leverage accessible labels or rewards to generate and filter reasoning steps, selecting those that yield correct solutions or high rewards~\cite{zelikman2022star,yuan2024scaling,singh2024beyond,hosseini2024vstar}. However, a critical limitation emerges from LLMs' tendency to generate incorrect or superfluous intermediate reasoning steps while still arriving at correct solutions through chance~\cite{debjit2024making}. This phenomenon significantly constrains the effectiveness of LLM fine-tuning for complex reasoning tasks~\cite{xia2024less,zhou2023lima}.

\section{Methodology}

\subsection{Notation}
Any reasoning task can be decomposed into three components: the problem tokens, the reasoning tokens (i.e., chain-of-thought steps), and the answer tokens. Let the problem token sequence be represented as \( \bm{x} = [x_1, x_2, \ldots, x_n] \). The Chain-of-Thought (CoT) process generates a sequence of intermediate reasoning tokens \( \bm{z} = [z_1, z_2, \ldots, z_m] \), where \( n \) and \( m \) denote the number of problem tokens and reasoning tokens respectively. In this work, we focus on a simple setting where the problem's answer is represented by a single token, denoted as \( y \). 

\textbf{CoT Auto-regressive Generation. } For auto-regressive generation, the mapping from the problem sequence \( \bm{x} \) to the answer \( y \) is performed through generating the intermediate tokens \( \bm{z} \) token by token. Formally, this can be expressed as:
\begin{align}
z_i \sim P(z_i|\bm{z_{<i}}, \bm{x}; \theta), \quad \text{for } i = 1, 2, \ldots, m,
\end{align}
where $\bm{z_{<i}}=[z_1, z_2, \ldots, z_{i-1}]$ represents precedent reasoning tokens. and the final answer is obtained at the final step:
\begin{align}
y \sim P(y|\bm{z},\bm{x};\theta).
\end{align}

\textbf{Looped Model. } Different from the auto-regressive model that generates explicit tokens to obtain the answer, the looped model implicitly maps the input sequence \( \bm{x} \) to the final answer \( y \) by executing the same function (e.g., a multi-layer Transformer block) for \( T \) times in the representation space. The number of iterations \( T \) depends on the problem comlexity. The forward process consists of three steps:
First, the token sequence \(\bm{x}\) is mapped to embeddings through an embedding function \(h\):
\begin{align}
\bm{e}_0 = h(\bm{x}; \theta_{\text{emb}}),
\end{align}
where \(\bm{e}_0 \in \mathbb{R}^{d\times n}\) and \(d\) is the hidden dimension. Second, the embeddings are iteratively refined through  transformation \(f\):
\begin{align}
\bm{e}_t = f(\bm{e}_{t-1}; \theta_{\text{model}}), \quad \text{for } t = 1, 2, \ldots, T,
\end{align}
where $f$ is usually a Transformer model and the number of iterations \(T\) is adaptively determined based on the problem length. Finally, the answer is predicted through a final-answer prediction head based on the representations in the last layer:
\begin{align}
y \sim P(y|\bm{e}_T; \theta_{\text{pred}}).
\end{align}
This design enables the model to perform implicit reasoning through iterative refinement in the representation space, where each iteration can automatically capture different aspects of the reasoning process. 

As a comparison, the CoT auto-regressive generation derives the final output \( y \) by first generating a sequence of intermediate reasoning tokens \( \bm{z} = [z_1, z_2, \ldots, z_m] \) in an auto-regressive manner, where the length $m$ can grow polynomially with input length \(n\) (i.e., \(m \sim \text{poly}(n)\)). This variable and potentially large \(m\) poses challenges for positional encoding to correctly reflect attention relationships, leading to low accuracy for long sequence reasoning. In contrast, the looped model takes a fundamentally different approach. It directly processes the input sequence \(\bm{x}\) and produces output \(y\). The network only needs to handle \(\bm{x}\) without \( \bm{z}\), mitigating the long-length problem in the reasoning chain.

\subsection{Length Generalization on Single Reasoning Task}\label{sec:single_task}
Before introducing our RELAY framework, we first empirically demonstrate the superior length generalization capability of looped Transformers compared to standard auto-regressive models. This  analysis serves as the foundation and motivation for our proposed framework.

\begin{figure*}[htbp]
    \centering
    \includegraphics[width=.9\textwidth]{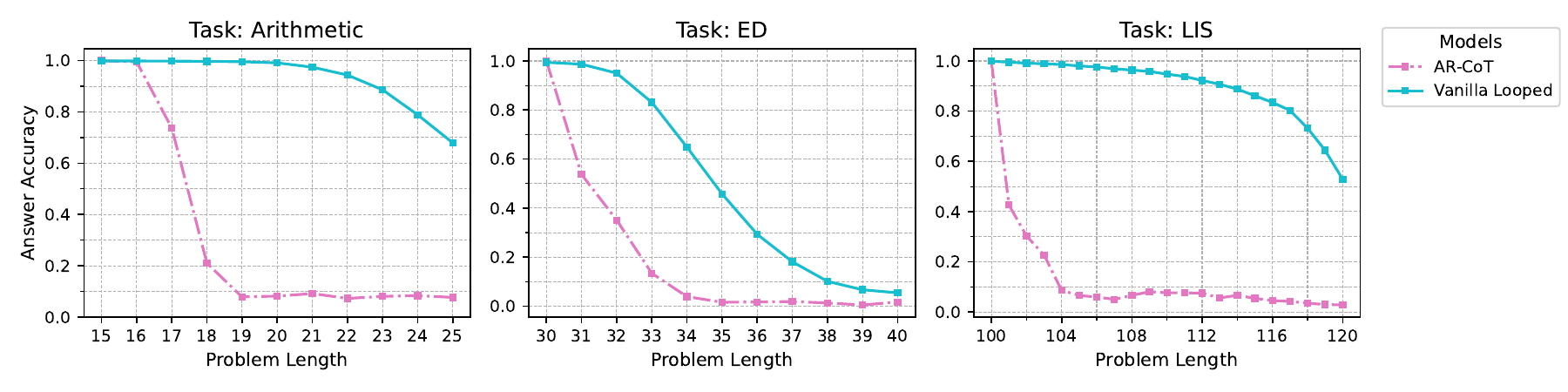}
    \vspace{-18pt}
    \caption{Length generalization performance of looped Transformer versus auto-regressive CoT model on Arithmetic (train: $\leq15$, test: $[15, 25]$), Edit Distance (train: $\leq30$, test: $[30, 40]$), and Longest Increasing Subsequence (train: $\leq100$, test: $[100, 120]$). } 
    \label{fig:single_tasks}
\end{figure*}

\textbf{Task Descriptions.} To validate the capabilities of different methods, we use three representative tasks adapted from~\citet{feng2024towards}, including Arithmetic, a mathematical task, and two dynamic programming (DP) problems: Edit Distance (ED) and Longest Increasing Subsequence (LIS). These tasks are selected for their diverse problem-solving patterns and varying levels of complexity, and the fact that they can be solved through a Chain-of-Thought reasoning process to arrive at the final answer. Performance is evaluated based on the accuracy of the final answer for both models. Detailed descriptions of these tasks are provided in Appendix~\ref{appendix:task_descriptions}.

\textbf{Experimental Setup.}
For each task, we construct a dataset consisting of 1 million training samples and 100\,k test samples, respectively. For the Arithmetic task, the problem complexity is defined as the number of operators. For the Edit Distance (ED) task, the problem complexity corresponds to the length of the shorter string in each pair. For the Longest Increasing Subsequence (LIS) task, we define the problem complexity as $\lceil n / 10 \rceil$, where $n$ is the length of the input sequence, as our dataset is structured with 10 numbers per reasoning step (see Appendix~\ref{appendix:task_descriptions} for details). The training datasets are constructed with the length of the problem token sequence  \(\bm{x}\) $\leq$ 15, 30, and 100 for Arithmetic, ED, and LIS, respectively. To evaluate the model's generalization capabilities, test datasets are created with problem lengths in the ranges $[15, 25]$ for Arithmetic, $[30, 40]$ for ED, and $[100, 120]$ for LIS. 



For the auto-regressive CoT model, we employ a standard decoder-only Transformer language model. For the looped model, we use an encoder-only Transformer with bi-directional attention. To address varying problem complexities, we implement dynamic iteration control in the looped model, setting the number of loop iterations equal to the problem complexity. The architectural configuration remains consistent across all models, comprising 3 layers, 256-dimensional hidden states, and 4 attention heads. For positional encoding, we adopt RoPE~\cite{su2024rope} across all model variants to enhance sequence encoding for both training and test cases.

\textbf{Results.}
Figure~\ref{fig:single_tasks} illustrates the comparative performance of both models. Within the training distribution (e.g. Arithmetic: $\leq15$ operators), both the looped Transformer and the auto-regressive CoT Transformer achieve perfect accuracy. However, the models exhibit markedly different behaviors when tested on problems exceeding the training length: While the auto-regressive CoT Transformer's performance deteriorates significantly, the looped Transformer maintains superior performance across all length regimes. This demonstrates the length generalization capabilities of the looped Transformer.

While looped Transformers exhibit superior performance in final answer prediction, they lack interpretability in their intermediate computational processes. Moreover, their design philosophy may struggle with general language tasks, as determining the number of loop iterations becomes challenging beyond reasoning problems. This work seeks to harness the accurate reasoning predictions of the looped model to guide the training of auto-regressive Chain-of-Thought (CoT) models to better handle long-sequence reasoning.

\begin{figure*}[t]
    \centering
    \includegraphics[width=\linewidth]{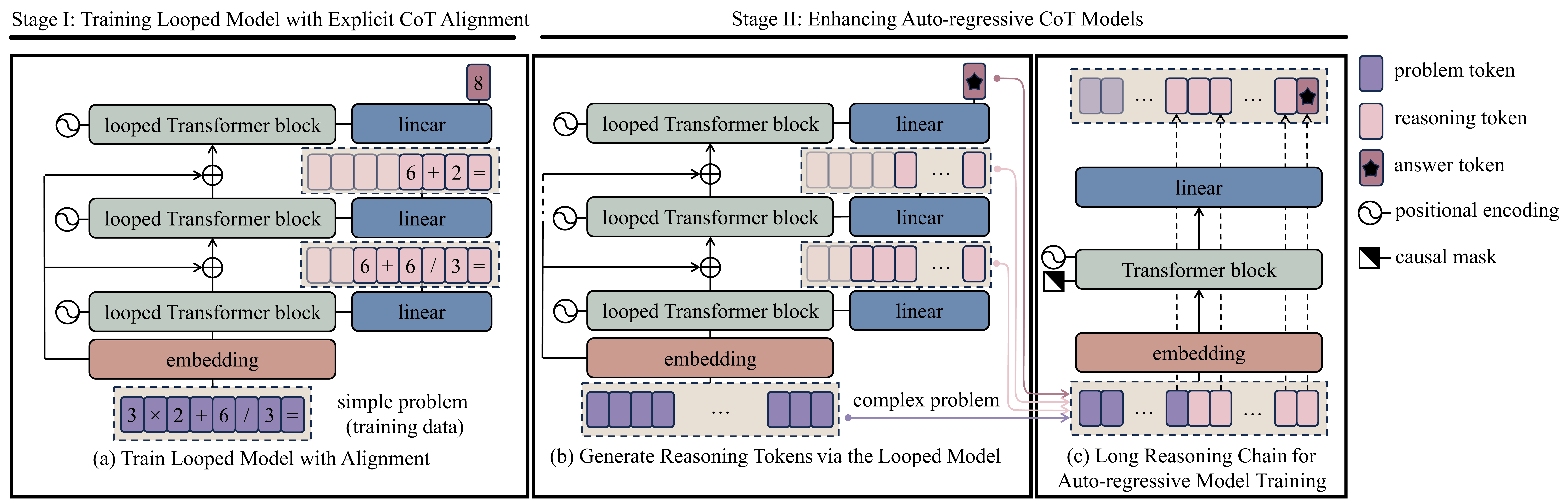}
    \caption{Overview of the RELAY framework. \textbf{Stage I (left):} Training looped model with explicit CoT alignment, where each iteration of the looped model learns to predict corresponding Chain-of-Thought (CoT) steps. \textbf{Stage II (right):} Using the trained looped model to generate CoT chains for enhancing auto-regressive CoT models. The looped model generates high-quality CoT chains for complex problems (beyond training length), which are then used to fine-tune the auto-regressive model to improve its reasoning capabilities.}
    \label{fig:loop_align_cot}
\end{figure*}

\subsection{Loop-Enhanced Chain-of-Thought Reasoning}

A straightforward way to leverage a well-trained looped model to enhance the auto-regressive CoT model is by using it as a verifier. When a problem is presented, both models generate a final answer, and if both answers match, the CoT output is trusted. However, this approach often fails in practice, as CoT models can produce incorrect reasoning trajectories even when reaching the correct final answer (see Section~\ref{ssec:relay_step_reliability}), making it unreliable to rely solely on the accuracy of the final answer as the guiding signal.



Our key insight is that an alignment can be established between the iterative structure of the looped Transformer and the stepwise nature of CoT reasoning. As shown in Figure~\ref{fig:loop_align_cot}, unlike the step-by-step token generation in CoT, looped models update their representations simultaneously in each iteration, and the number of such iterations naturally corresponds to the number of reasoning rounds. This structural similarity opens up the possibility of training the looped model to generate the corresponding CoT tokens for each round in parallel, while maintaining its ability to predict the final answer. With this insight, we propose \textbf{RELAY} (\underline{\textbf{RE}}asoning through \underline{\textbf{L}}oop \underline{\textbf{A}}lignment iterativel\underline{\textbf{Y}}), a two-stage framework that bridges looped and auto-regressive models.

\textbf{Stage I: Training Looped Model with Explicit CoT Alignment.}
In the first stage, we train the looped model to generate intermediate reasoning processes that align with CoT steps. To formalize this alignment, assume we have a reasoning chain with $T$ rounds. Given reasoning tokens $\bm{z}=[z_1, z_2, \ldots, z_m]$, denote $k_t$ as the start token position of $t$-th reasoning round, where each round contains valid reasoning tokens $\bm{z}_{[k_t:k_{t+1}-1]}=[z_{k_t}, z_{k_{t+1}}, \ldots, z_{k_{t+1}-1}]$.  Taking arithmetic reasoning as an example, consider a sequence of tokens representing the complete reasoning chain, ``$3\times 2 + 6 \div 3 = 6 + 6 \div 3 = 6 + 2 = 8$''. This sequence can be naturally divided into $T=3$ rounds using the equal signs as delimiters: Given the input problem ``$3\times 2 + 6 \div 3 =$'', the first round corresponds to ``$6 + 6 \div 3 = $'', the second round corresponds to ``$6 + 2 = $'', and the third (last) round presents the final answer ``$8$''. 


Although the number of rounds aligns with the iteration count of the looped model, a key challenge arises from the mismatch in token lengths across different reasoning steps. For instance, earlier steps involving complex expressions (e.g., ``$6 + 6 \div 3 = $'') typically require more tokens than later steps (e.g., ``$6 + 2 = $''). This variable length nature poses a challenge for the looped model, which requires fixed-length representations of size $n$ across iterations.

To address this length mismatch while preserving the parallel processing capability of the looped model, we employ a right-aligned padding strategy. For the $t$-th iteration, we construct a fixed-length sequence $\tilde{\bm{z}}_t$ of length $n$ by right-aligning the ground truth reasoning tokens $\bm{z}_{[k_t:k_{t+1}-1]}$ and filling the remaining left positions with \texttt{<pad>} tokens. The fixed-length is determined based on the maximum length among all reasoning rounds and the original input problem (note that the length of a reasoning round usually does not exceed the length of the input problem; otherwise, each round can be further divided into shorter rounds). 
To track both valid reasoning tokens and the boundary of padding, we introduce a binary mask:
\begin{equation}
M_t[i] = \begin{cases}
1, & \text{if } i = p_t \text{ or } \tilde{\bm{z}}_t[i] \neq \text{\texttt{<pad>}}, \\
0, & \text{otherwise},
\end{cases}
\end{equation}
where $M_t$ indicates the positions of valid reasoning tokens and the position of the last \texttt{<pad>} token $p_t$.

Using this alignment strategy, we train the looped model to predict the corresponding CoT tokens at each iteration, enabling it to generate CoT-aligned intermediate outputs. In detail, at each iteration $t$, we train the model to predict both the valid reasoning tokens and the last \texttt{<pad>} token through an intermediate prediction head:
\begin{equation}
P(\tilde{\bm{z}}_t|\bm e_t; \theta_{\text{pred-cot}}),
\end{equation}
For the intermediate reasoning steps, we ignore all preceding \texttt{<pad>} tokens except the last one, as they have no impact on the reasoning process. The loss of this part can be formulated as :
\begin{equation}
\mathcal{L}_{\text{iter}} = \frac{1}{T}\sum_{t=1}^T \text{CrossEntropy}(P(\tilde{\bm{z}}_t|\bm e_t;\theta_{\text{pred-cot}}), \tilde{\bm{z}}_t) \odot M_t,
\end{equation}
where the element-wise multiplication $\odot$ ensures that the loss is computed only on valid reasoning tokens and the last \texttt{<pad>} token.

For the final answer, we have the answer prediction loss to ensure correct final predictions:
\begin{equation}
\mathcal{L}_{\text{ans}} = \text{CrossEntropy}(P(\bm{y}|\bm{e}_T;\theta_{\text{pred}}), \bm{y}),
\end{equation}
where $\bm{y}$ is the ground truth answer.
The total training loss is then:
\begin{equation}
\mathcal{L} = \mathcal{L}_{\text{ans}} + \lambda\mathcal{L}_{\text{iter}},
\end{equation}
where $\lambda$ is a hyperparameter balancing the two objectives.

This design enables the looped model to accurately predict the answer and provide interpretable intermediate reasoning steps that can be effectively utilized to guide the auto-regressive model in Stage II.

\paragraph{Stage II: Enhancing Auto-regressive CoT Models.}
In the second stage, we leverage the trained looped model to enhance auto-regressive CoT models through a systematic process:

First, we use the trained looped model in Stage I to generate reasoning demonstrations for problems of increasing complexity. For each problem $x$, we obtain:
\begin{equation}
(\bm{z}, y) \sim p(\cdot|\bm{x};\theta_L),
\end{equation}
where $\theta_L$ denotes the trained looped model from Stage I, $\bm{z} = [z_1, z_2, \ldots, z_m]$ represents the generated reasoning tokens across iterations, and $y$ is the predicted answer.



We then utilize these demonstrations to fine-tune an auto-regressive model. For problem lengths beyond the original training range, we generate a comprehensive dataset of reasoning demonstrations using the looped model. This newly generated data is then merged with the original training dataset, which contains problems within the initial training length. The combined dataset, spanning both the original and extended problem lengths, is then used to fine-tune the auto-regressive model in a single step. This approach allows the model to retain its original reasoning capabilities while acquiring the ability to effectively tackle more complex, longer problems, utilizing the structured insights provided by the demonstrations.

\textbf{Comparison with Synthetic Data Generation Approaches. }
To effectively guide the LLMs to handle complex problems, prior works~\cite{hendrycks2021measuring,lightman2024lets} have explored the synthetic data generation approach, where human labelers construct data generation pipelines based on their understanding of both the task and its solution process. This approach requires labelers to possess comprehensive knowledge in three aspects: (1) problem construction, (2) problem-solving strategies, and (3) pipeline development skills. While effective, this creates a high barrier for deployment across diverse domains, as finding experts who excel in all three areas can be challenging.

In contrast, RELAY  reduces these requirements. Our approach follows a more automated pipeline: training data → looped model with strong generalization capability → longer problem construction → automated reasoning generation → auto-regressive CoT model training. The human involvement is primarily limited to longer problem construction, eliminating the need for expertise in solution strategies and pipeline development. This reduction in human expertise requirements makes our method more practical and scalable across different domains. Additionally, by leveraging the looped models' inherent generalization capabilities rather than manually designed rules, our approach can potentially capture more nuanced reasoning patterns that might be overlooked in hand-crafted pipelines.

\section{Experiments}\label{sec:exp}

This section presents a comprehensive empirical evaluation of our RELAY framework through a series of experiments designed to address four key research questions:
\vspace{-6pt}
\begin{itemize}
\item \textbf{Q1}: How effectively does the looped model with explicit CoT alignment serve as a general-purpose reasoner across diverse tasks?
(Section~\ref{sec:multitask})
\item \textbf{Q2}: What advantages does the looped model with explicit CoT alignment demonstrate in length generalization compared to auto-regressive CoT models?
(Section~\ref{sec:multitask})
\item \textbf{Q3}: How can the length generalization capabilities of the looped model with explicit CoT alignment be leveraged to enhance auto-regressive CoT models?
(Section~\ref{ssec:relay_enhance_cot})
\item \textbf{Q4}: How reliable are the intermediate reasoning steps generated by the looped model with explicit CoT alignment?
(Section~\ref{ssec:relay_step_reliability})
\end{itemize}
\vspace{-6pt}
We address each question through carefully designed experiments, as detailed below.

\begin{figure*}[t]
    \centering
    \includegraphics[width=0.95\textwidth]{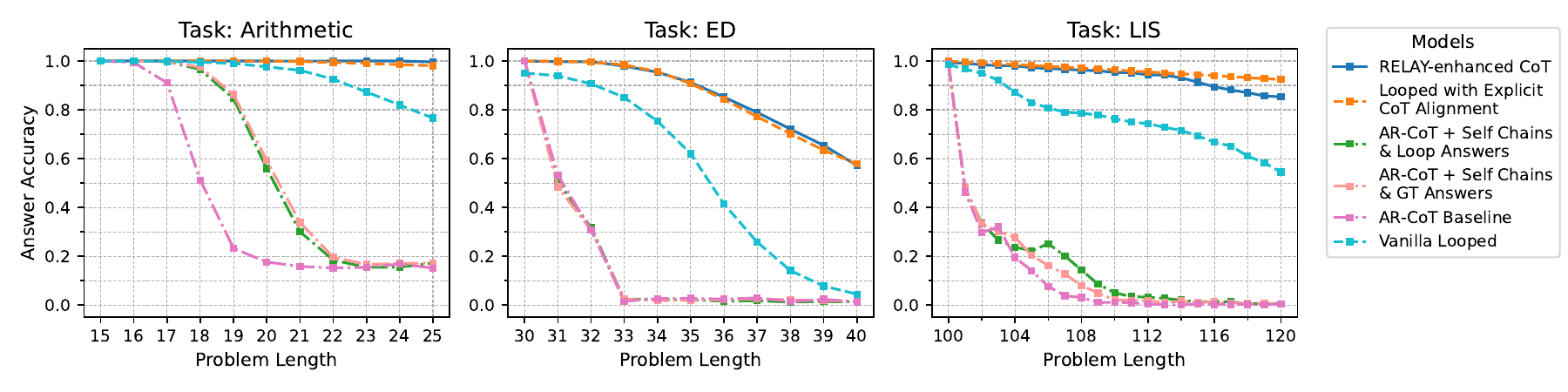}
    \vspace{-16pt}
    \caption{Performance comparison of different models on long reasoning problems across three tasks: Arithmetic, Edit Distance (ED), and Longest Increasing Subsequence (LIS).}
    \label{fig:main_result}
\end{figure*}

\subsection{Multitask Training}\label{sec:multitask}
Following the single-task evaluation discussed in Section~\ref{sec:single_task}, we extend our analysis to a multitask learning setting to explore the general reasoning capabilities of three models: the looped model with explicit CoT alignment, the auto-regressive CoT model, and the vanilla looped model. In this setup, we jointly train the models on three representative reasoning tasks: Arithmetic, Edit Distance (ED), and Longest Increasing Subsequence (LIS), each requiring multi-step reasoning to arrive at accurate final answers. This setting enables a thorough comparison of the models' ability to generalize effectively across diverse tasks.

For a detailed description of the tasks, including example inputs, expected answers, and the corresponding Chain-of-Thought (CoT) reasoning steps, please refer to Appendix~\ref{appendix:task_descriptions}.




\textbf{Experimental Setup. }
We conduct experiments on the three tasks: Arithmetic, Edit Distance (ED), and Longest Increasing Subsequence (LIS), to evaluate the generalization capabilities of the looped model with explicit CoT alignment in comparison with the auto-regressive CoT model and the vanilla looped model. Training datasets 
retain the same problem lengths as in Section~\ref{sec:single_task}: operator counts of $\leq 15$ for Arithmetic, input string lengths of $\leq 30$ for ED, and sequence lengths $\leq 100$ for LIS. Similarly, test datasets are constructed with extended problem lengths of $[15, 25]$, $[30, 40]$, and $[100, 120]$ for Arithmetic, ED, and LIS, respectively, to assess length generalization.

In this setup, each model---the looped model with explicit CoT alignment, the auto-regressive CoT model, and the vanilla looped model---is trained jointly on all three tasks by prepending a task-specific problem token (\texttt{[ARI]}, \texttt{[ED]}, \texttt{[LIS]}) to the input sequence, which distinguishes among tasks. All models are evaluated under the same metric, considering only the accuracy of final answer.





\textbf{Results. }
Figure~\ref{fig:main_result} illustrates the comparative performance of the three models across the three tasks. All models achieve nearly 100\,\% accuracy on all tasks within the training distribution, demonstrating that the looped model with explicit CoT alignment can serve as a general-purpose reasoning engine capable of handling diverse tasks requiring multi-step reasoning. (\textbf{Q1})

However, for problems with lengths exceeding the training range, the looped model with explicit CoT alignment and the vanilla looped model significantly outperform the auto-regressive CoT model, showcasing the superiority of loop-based architectures in addressing tasks requiring generalization to longer inputs. 
Furthermore, the looped model with explicit CoT alignment not only maintains the strong length generalization capability of the vanilla looped model but even surpasses it notably, benefiting from the explicit alignment between CoT reasoning steps and loop iterations. This alignment provides structural guidance that enhances the model's reasoning capabilities over extended lengths as well as the ability to generate explicit intermediate CoT reasoning chains, making it both accurate and interpretable. 
These results establish the looped model with explicit CoT alignment as both a robust reasoning framework and a generally effective solution for length generalization challenges, outperforming standard auto-regressive CoT models across diverse tasks. (\textbf{Q2})

\subsection{Enhancing Auto-regressive Model with RELAY-Generated CoT Data}\label{ssec:relay_enhance_cot}

In this section, we utilize the looped model with explicit CoT alignment trained in Section~\ref{sec:multitask} to enhance the performance of the auto-regressive CoT model via effective data generation. Specifically, we leverage its ability to produce accurate reasoning chains for complex problems exceeding the training lengths. These reasoning chains serve as high-quality data, which are subsequently employed to fine-tune the auto-regressive CoT model.

\textbf{Experimental Setup. }
First, we employ the looped model with explicit CoT alignment to generate CoT reasoning chains for problems of increased complexity, covering problem lengths of $[15, 25]$, $[30, 40]$, and $[100, 120]$ for Arithmetic, ED, and LIS tasks, respectively. These newly generated data is then merged with the original training dataset, which contains problems within the initial training length. Details of the sample proportions for different problem lengths when merging datasets are provided in Appendix~\ref{appendix:samples_dist_dataset}.

Next, we fine-tune the auto-regressive CoT model on this augmented dataset in a single phase. This fine-tuning process builds upon the well-trained auto-regressive CoT model from Section~\ref{sec:multitask}, retaining the same model structure
while updating the weights with the augmented dataset. This process enables the model to incorporate longer CoT chains, thereby enhancing its reasoning capabilities on extended sequences.

\textbf{Results.} Figure~\ref{fig:main_result} presents the accuracy curves of the 
RELAY-enhanced auto-regressive CoT model 
across problem lengths for the three tasks. Compared to the baseline auto-regressive CoT model, the auto-regressive CoT model fine-tuned with data generated by RELAY (i.e., by the looped model with explicit CoT alignment) exhibits significant improvements on problems exceeding the original training length. Notably, its performance approaches and even slightly surpasses that of the looped model with explicit CoT alignment in some cases, while consistently outperforming the baseline auto-regressive CoT model.

These results indicate that our RELAY framework effectively utilizes the length generalization capabilities of the looped model with explicit CoT alignment to improve the overall performance of auto-regressive model. By generating high-quality CoT reasoning data, RELAY enables the auto-regressive CoT model to better handle problems beyond its original training range, without altering its architecture. (\textbf{Q3})

\subsection{Evaluating the Reliability of RELAY-Generated Intermediate Reasoning Steps}\label{ssec:relay_step_reliability}
This section aims to demonstrate the reliability of CoT chains generated by the looped model with explicit CoT alignment compared to the auto-regressive CoT model's self-generated data. Specifically, we highlight that the generated data from the auto-regressive CoT model, even when the final result is correct, often contains incorrect intermediate steps, which is why utilizing these data fails to improve the model's performance in complex problems with longer lengths. In contrast, data generated by the looped model with explicit CoT alignment avoids these issues by ensuring both accurate intermediate reasoning steps and the final answer, enabling effective fine-tuning of the auto-regressive model and significantly enhancing its performance.

\begin{figure}
    \centering
    \includegraphics[width=.5\textwidth]{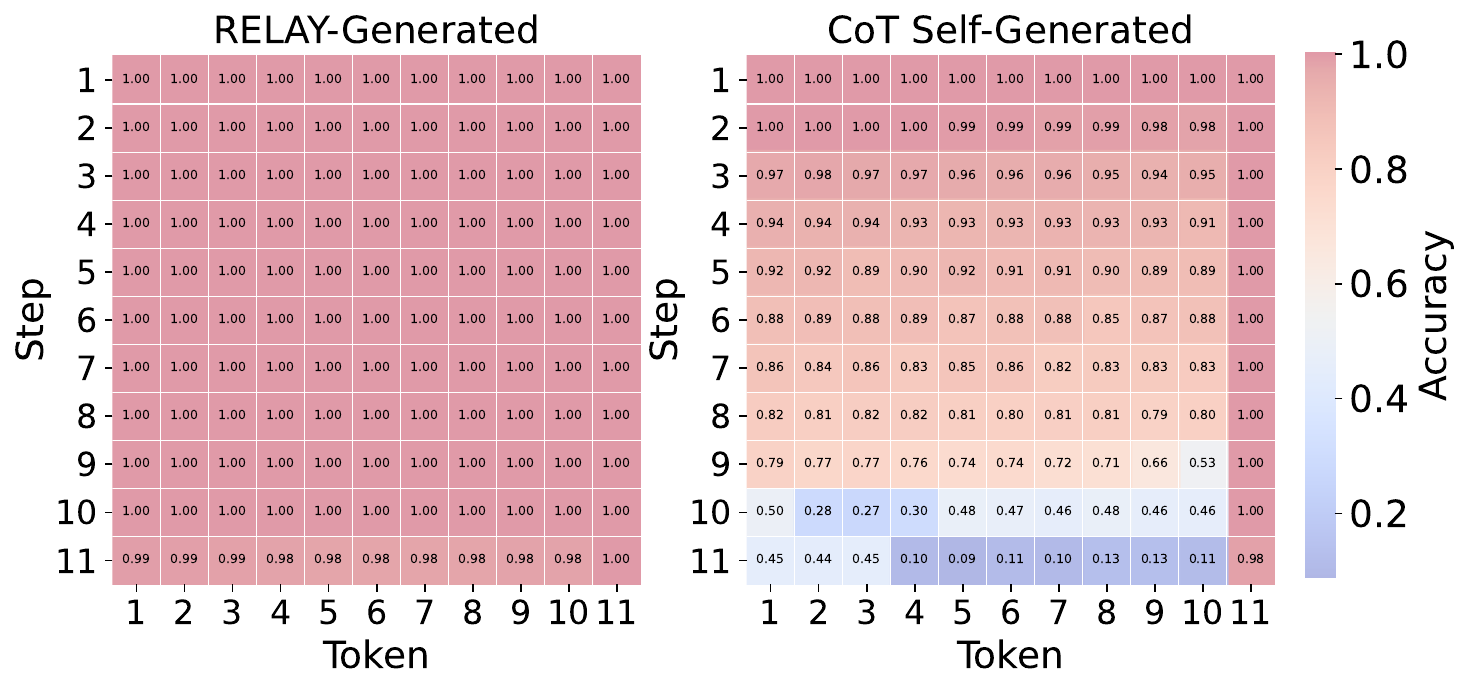}
    \vspace{-20pt}
    \caption{Hit accuracy matrices for the LIS task with a problem length of 105 ($T = 11$ steps).} 
    \label{fig:matrix}
    \vspace{-5pt}
\end{figure}


\textbf{Experimental Setup. }
We evaluate the effectiveness of two types of generated data: by the looped model with explicit CoT alignment and the auto-regressive CoT model. This evaluation focuses on two metrics: (1) hit matrix and (2) bit accuracy, which provides a detailed perspective on reasoning steps reliability.

For the hit matrix, we select the LIS task as an example due to the structured nature of its reasoning steps. The intermediate reasoning steps of LIS tasks follows a $T\times 11$ matrix format, where $T$ corresponds to the number of CoT steps as well as the iteration number of the looped model, and 11 represents the number of tokens per step (10 numbers as prescribed in our dataset, along with one delimeter \texttt{<sep>}). This structured format makes the LIS task particularly suitable for evaluating and visualizing reasoning step reliability, offering an intuitive representation of the proportion of tokens at each position that match the ground truth reasoning steps.

Bit accuracy is provided across all three tasks of varying lengths, evaluating token-wise counted accuracy for the whole reasoning step. Comparisons are made between the auto-regressive CoT models fine-tuned by the two types of generated data respectively. 



For the auto-regressive CoT model self-generated data, we conduct the following experiment under two parallel settings, using either a vanilla looped model or ground-truth answers as verifiers. The experiment consists of the following steps: (1) Use the auto-regressive CoT model to generate CoT chains for long problem lengths. (2) Filter these data by the looped model or ground-truth, retaining only those where the final answers match. (3) The filtered data are then used to fine-tune the auto-regressive CoT model. The fine-tuning process aims to improve the model's ability to generate reasoning trajectories and reach the correct final answer for longer problems. 

Meanwhile, data generated by the looped model with explicit CoT alignment is employed as the fine-tuning dataset for the same initial auto-regressive CoT model checkpoint, under the same fine-tuning parameters and controlled ratio of samples with different lengths. Both approaches are evaluated across three tasks with varying lengths to assess performance improvements.

\textbf{Results.} We evaluate the hit accuracy matrix for the LIS task with a problem length of 105, which corresponds to $T = \lceil 105 / 10 \rceil = 11$ steps, resulting in an $11 \times 11$ matrix (We also provide results for problem length of 101 in Appendix~\ref{appendix:hit_matrix_101}). As shown in Figure~\ref{fig:matrix}, data generated by the looped model with explicit CoT alignment achieves consistently high token accuracy across all positions, with most values approaching 100\,\%, demonstrating its ability to produce high-quality and reliable data. (\textbf{Q4}) In contrast, the data generated by the auto-regressive CoT model exhibits high token accuracy only in the first few positions, while the accuracy steadily decreases in later steps. Although the delimiter tokens \texttt{<sep>} at the end of each step  achieve high accuracy, this simply implies that the auto-regressive CoT model has only captured the basic format of the reasoning process but fails to predict accurate tokens, which indicates its limited capability to maintain accurate prediction throughout the reasoning process for longer problems.

The bit accuracy results for the models fine-tuned with different datasets across the three tasks (Arithmetic, ED, LIS) and varying problem lengths are provided in Appendix~\ref{appendix:bit_acc_results}. 

We additionally provide the accuracy of the final answer for the auto-regressive CoT model fine-tuned with self-generated data in Figure~\ref{fig:main_result}, noted as ``AR-CoT + Self Chains \& Loop/GT Answers'', corresponding to data filtered by the looped model or ground-truth answers, respectively, which only shows a slight improvement over the baseline model.

\section{Conclusion}

This paper introduces \textbf{RELAY} (\underline{\textbf{RE}}asoning through \underline{\textbf{L}}oop \underline{\textbf{A}}lignment iterativel\underline{\textbf{Y}}), a framework enhancing Chain-of-Thought reasoning by combining looped and auto-regressive Transformers. Our contributions show that (1) a looped Transformer can serve as a general-purpose reasoner with strong length generalization, (2) iteration-wise alignment enables accurate reasoning chain generation beyond training length, and (3) RELAY improves auto-regressive models through high-quality generated reasoning chains. Future work could explore the theoretical foundations of looped Transformers' length generalization and extend RELAY to broader language tasks.



\newpage

\section*{Impact Statement}

This paper presents work whose goal is to advance the field of Machine Learning. There are many potential societal consequences of our work, none which we feel must be specifically highlighted here.

\bibliography{main}
\bibliographystyle{icml2025}

\newpage
\appendix
\onecolumn

\section{Task Descriptions}\label{appendix:task_descriptions}
Below, we present the detailed descriptions of each task from~\citet{feng2024towards}, including examples of inputs, expected answers, and the corresponding Chain-of-Thought (CoT) reasoning steps used to derive the final answers.

\begin{enumerate}
    \item \textbf{Arithmetic}. This task involves computing the answer of arithmetic expressions containing numbers, basic operations ($+,-,\times,\div,=$), and brackets. For example:
    \begin{itemize}
        \item \textbf{Input:} $(6+9)\div(7+2\times 5-4\times 3) = $
        \item \textbf{CoT Steps:}
        \begin{align*}
        & 15\div(7+2\times 5-4\times 3) =  \\
        & 15\div(7+10-4\times 3) =  \\
        & 15\div(17-4\times 3) =  \\
        & 15\div(17-12) =  \\
        & 15\div 5 = 
        \end{align*}
        \item \textbf{Answer:} 3
    \end{itemize}

    \item \textbf{Edit Distance (ED)}. This task requires computing the minimum number of operations (insert, delete, or replace) needed to transform one sequence into another.
    The input consists of two sequences separated by a delimiter \texttt{|}:
    \begin{itemize}
        \item \textbf{Input:} \texttt{o t m l | o t t m l <sep>}
        \item \textbf{CoT Steps:}
        \begin{verbatim}
        0 2 4 6 7 ,
        2 0 2 4 6 ,
        4 2 3 2 4 ,
        6 4 5 4 2 ,
        \end{verbatim}
        \item \textbf{Answer:} \texttt{2}
    \end{itemize}
    Each row corresponds to the edit distance matrix, and the final answer is the edit distance.
    
    \item \textbf{Longest Increasing Subsequence (LIS)}. This task identifies the length of longest strictly increasing subsequence in a numerical sequence.
    The input is a sequence of integers followed by a delimiter \texttt{<sep>}:
    \begin{itemize}
        \item \textbf{Input:} \texttt{103 110 145 217 233 18 30 82 141 150 159 161 167 239 <sep>}
        \item \textbf{CoT Steps:}
        \begin{verbatim}
        1 2 3 4 5 1 2 3 4 5 <sep>
        6 7 8 9 9 9 9 9 9 9 <sep>
        \end{verbatim}
        \item \textbf{Answer:} \texttt{9}
    \end{itemize}
    Here, each CoT step represents an intermediate computation in the dynamic programming process, folded into fixed-size groups (10 numbers per step in our setting) to align with the model structure. If the last group has fewer than 10 numbers, the last number is repeated until the group size reaches 10.
\end{enumerate}

\section{Hit Matrix for LIS Task with Length 101}\label{appendix:hit_matrix_101}
\begin{figure*}[t]
    \centering
    \includegraphics[width=.8\textwidth]{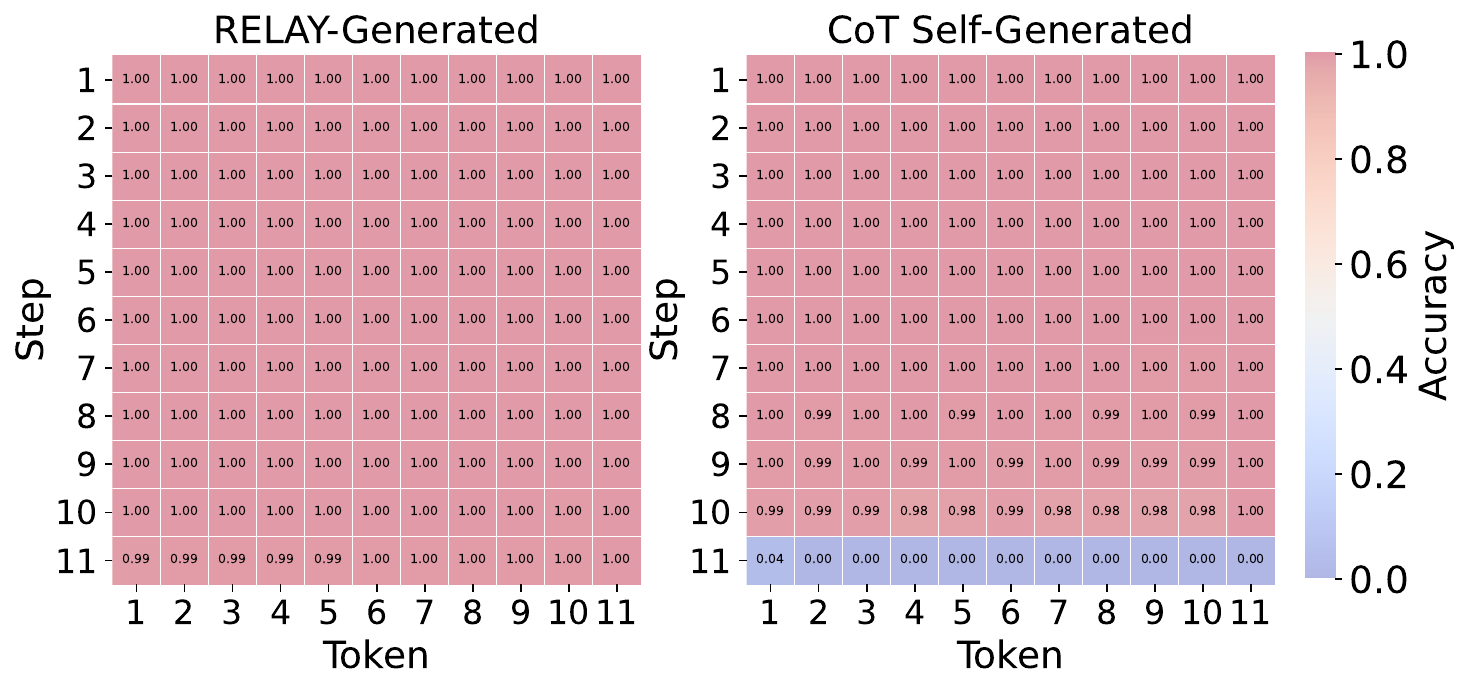}
    \vspace{-8pt}
    \caption{Hit accuracy matrices for the LIS task with a problem length of 101 ($T = 11$ steps).} 
    \label{fig:hit_mat_101}
\end{figure*}

We also analyze the hit accuracy matrix for the LIS task with a problem length of 101, corresponding to $T = \lceil 101 / 10 \rceil = 11$ reasoning steps, as shown in Figure~\ref{fig:hit_mat_101}. The results exhibit a similar trend to those observed for length 105, with a notable decline in accuracy in the later reasoning steps for data generated by the auto-regressive CoT model, while RELAY-generated data consistently maintains high accuracy. Specifically, while the initial steps maintain relatively high token accuracy, the accuracy deteriorates significantly in later steps, failing to achieve accurate final answers. This highlights a key limitation of using CoT self-generated data for supervision: even when the problem length is only slightly beyond the training length, the CoT model struggles to generate accurate reasoning steps towards the end, making it infeasible to fine-tune the model using only the final answer as supervision, as the lack of intermediate reasoning accuracy prevents meaningful improvements in model's performance of handling longer problems.

\section{Bit Accuracy}\label{appendix:bit_acc_results}

\begin{figure*}[htbp]
    \centering
    \includegraphics[width=.95\textwidth]{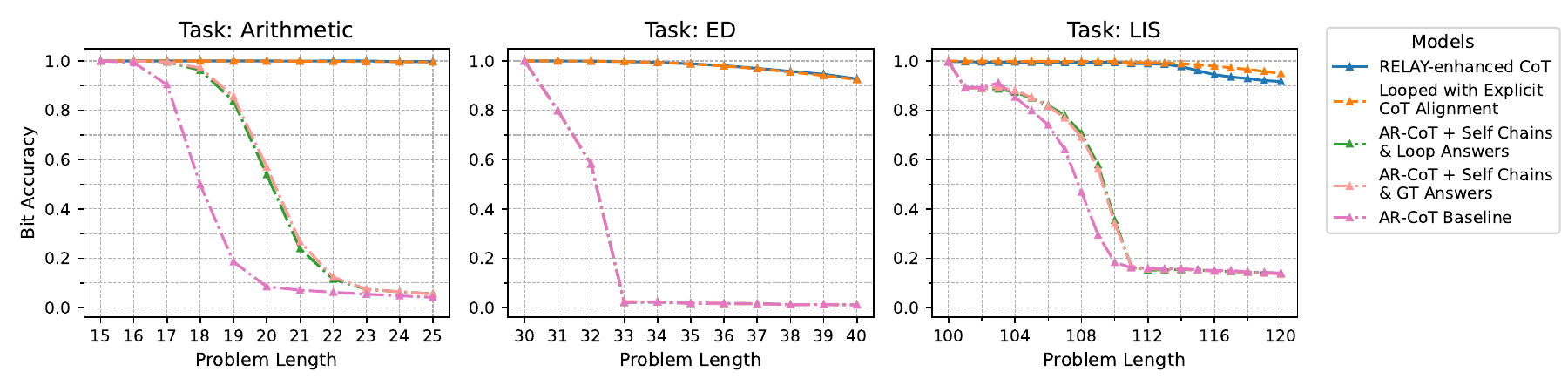}
    \vspace{-8pt}
    \caption{Bit accuracy over varying problem lengths for three tasks: Arithmetic, ED, and LIS.} 
    \label{fig:bit_acc}
\end{figure*}

The bit accuracy results for the models fine-tuned with different datasets across the three tasks (Arithmetic, ED, LIS) and varying problem lengths are shown in Figure~\ref{fig:bit_acc}. Each subfigure corresponds to one task and includes curves showing the bit accuracy of five models over varying problem lengths: (1) \textbf{RELAY-enhanced CoT:} the auto-regressive CoT model fine-tuned with data generated by RELAY, (2) \textbf{Looped with Explicit CoT Alignment}, (3) \textbf{AR-CoT + Self Chains
\& Loop Answers:} the auto-regressive CoT model fine-tuned with its self-generated data using looped model answers as labels, (4) \textbf{AR-CoT + Self Chains \& GT Answers:} the auto-regressive CoT model fine-tuned with its self-generated data using ground-truth answers as labels, and (5) \textbf{AR-CoT Baseline:} the baseline auto-regressive CoT model as a reference, also as the initial auto-regressive CoT model before fine-tuned.

As illustrated in Figure~\ref{fig:bit_acc}, both the looped model with CoT alignment and the auto-regressive CoT model fine-tuned with data generated by it consistently achieve high bit accuracy across all tasks and problem lengths. Notably, they maintain over 90\,\% bit accuracy even at lengths extending beyond the training data (up to $+10$ for Arithmetic and ED tasks, and $+20$ for the LIS task in our setting). This highlights not only the robustness and reliability of the looped model with CoT alignment in length extrapolation scenarios but also the effectiveness of its generated data in significantly enhancing the performance of the auto-regressive CoT model.

In contrast, the auto-regressive CoT model fine-tuned with its own self-generated data shows limited improvement over the baseline model. This is consistent across all tasks and highlights a critical limitation: the self-generated data often contain incorrect intermediate steps, even when the final results are correct. These inaccuracies hinder the model's ability to generalize and perform well on longer problem lengths, reinforcing the importance of reliable intermediate reasoning steps for effective fine-tuning.

\section{Details for training and fine-tuning}\label{appendix:details_train_finetune}
\subsection{Hyper-parameters}\label{appendix:hyperparameters}

In our experiments, we trained three different models: (1) the looped model with CoT alignment, (2) the auto-regressive CoT model, and (3) the vanilla looped model. All models were trained from scratch on the same dataset, which consists of 1 million samples for each of the three tasks. The task-specific training weights and training hyper-parameters are provided in Table~\ref{tab:training_hyperparams}. 

\begin{table}[htbp]
  \centering
  \caption{Training Hyper-parameters of Different Models}
  \vskip 0.15in
    \begin{tabular}{cccc}
    \toprule
    Training Hyper-parameters & Looped Model with CoT Alignment & CoT Model & Vanilla Looped Model \\
    \midrule
    Epoch & 500   & 500   & 500 \\
    Batch Size & 512   & 512   & 512 \\
    Learning Rate & 5e-4 & 5e-4 & 1e-3 \\
    Learning Rate Schedule & linear & linear & linear \\
    Warmup Ratio & 0.01  & 0.01  & 0.01 \\
    Optimizer & AdamW & AdamW & AdamW \\
    Weight Decay & 0.01  & 0.01  & 0.01 \\
    Drop out & 0.1   & 0.1   & 0.1 \\
    Weight of ARI & 1     & 1     & 1 \\
    Weight of ED & 1     & 10    & 10 \\
    Weight of LIS & 1     & 5     & 5 \\
    \bottomrule
    \end{tabular}%
  \label{tab:training_hyperparams}%
\end{table}%

For fine-tuning, we used data generated by RELAY and CoT model self-generated data to fine-tune the auto-regressive CoT model. The fine-tuning process followed a similar setup, as detailed in Table~\ref{tab:finetune_hyperparams}. 

\begin{table}[htbp]
  \centering
  \caption{Fine-tuning Hyper-parameters}
  \vskip 0.15in
    \begin{tabular}{ccc}
    \toprule
    Fine-tuning Hyper-parameters & RELAY-Generated Data & Self-Generated Data \\
    \midrule
    Epoch & 500 & 100 (per phase) \\
    Batch Size & 512 & 512 \\
    Learning Rate & 1e-4 & 5e-5 \\
    Learning Rate Schedule & linear & linear \\
    Warmup Ratio & 0.01 & 0.01 \\
    Optimizer & AdamW & AdamW \\
    Weight Decay & 0.01 & 0.01 \\
    Drop out & 0.1 & 0.1 \\
    Weight of ARI & 1 & 1 \\
    Weight of ED & 10 & 1 \\
    Weight of LIS & 5 & 1 \\
    \bottomrule
    \end{tabular}%
  \label{tab:finetune_hyperparams}%
\end{table}%

\subsection{Sample Length Distribution of Datasets}\label{appendix:samples_dist_dataset}



The original training dataset for training the three models consists of 1 million samples for each task, following a distribution where the number of samples is proportional to problem length.

The merged dataset used for fine-tuning consists of 100\,k samples for each of the three tasks, incorporating both the original training data and newly generated samples from extended problem lengths. 

For the looped model with explicit CoT alignment, we introduce additional data covering problem lengths of $[16, 25]$ for Arithmetic, $[31, 40]$ for ED, and $[101, 120]$ for LIS. These newly generated samples are merged with the original dataset while maintaining a balanced proportion across different length ranges to ensure effective training. 
The specific numbers of samples for different problem lengths in the final merged dataset are provided in Table~\ref{tab:sample_dist_RELAY}.

\begin{table}[htbp]
  \centering
  \vspace{-8pt}
  \caption{Number of Samples for Different Problem Lengths in Merged Dataset}
  \vskip 0.15in
    \begin{tabular}{cccc}
    \toprule
    Task & Arithmetic & ED    & LIS \\
    \midrule
    Length & $\leq 15$ & $\leq 30$ & $\leq 100$ \\
    Number of Samples & 42515 & 60844 & 73235 \\
    \midrule
    Length & 16    & 31    & 101 \\
    Number of Samples & 6477  & 4195  & 1479 \\
    \midrule
    Length & 17    & 32    & 102 \\
    Number of Samples & 6882  & 4195  & 1464 \\
    \midrule
    Length & 18    & 33    & 103 \\
    Number of Samples & 7287  & 4055  & 1449 \\
    \midrule
    Length & 19    & 34    & 104 \\
    Number of Samples & 6477  & 4055  & 1434 \\
    \midrule
    Length & 20    & 35    & 105 \\
    Number of Samples & 6072  & 3916  & 1420 \\
    \midrule
    Length & 21    & 36    & 106 \\
    Number of Samples & 5668  & 3916  & 1405 \\
    \midrule
    Length & 22    & 37    & 107 \\
    Number of Samples & 5263  & 3776  & 1390 \\
    \midrule
    Length & 23    & 38    & 108 \\
    Number of Samples & 4858  & 3776  & 1375 \\
    \midrule
    Length & 24    & 39    & 109 \\
    Number of Samples & 4453  & 3636  & 1360 \\
    \midrule
    Length & 25    & 40    & 110 \\
    Number of Samples & 4048  & 3636  & 1346 \\
    \midrule
    Length &       &       & 111 \\
    Number of Samples &       &       & 1331 \\
    \midrule
    Length &       &       & 112 \\
    Number of Samples &       &       & 1316 \\
    \midrule
    Length &       &       & 113 \\
    Number of Samples &       &       & 1301 \\
    \midrule
    Length &       &       & 114 \\
    Number of Samples &       &       & 1286 \\
    \midrule
    Length &       &       & 115 \\
    Number of Samples &       &       & 1272 \\
    \midrule
    Length &       &       & 116 \\
    Number of Samples &       &       & 1257 \\
    \midrule
    Length &       &       & 117 \\
    Number of Samples &       &       & 1242 \\
    \midrule
    Length &       &       & 118 \\
    Number of Samples &       &       & 1227 \\
    \midrule
    Length &       &       & 119 \\
    Number of Samples &       &       & 1213 \\
    \midrule
    Length &       &       & 120 \\
    Number of Samples &       &       & 1198 \\
    \bottomrule
    \end{tabular}%
  \label{tab:sample_dist_RELAY}%
\end{table}%

For the self-generated dataset, we adopt an incremental approach, since the accuracy of original CoT model diminishes rapidly as the problem length increases. Specifically, we maintain a total dataset size of 100\,k samples for each task. The initial dataset consists of problems with lengths $\leq 15, 30,$ and 100 for Arithmetic, ED, and LIS, respectively. The CoT model is progressively fine-tuned over five phases, each including self-generation on slightly longer problems and followed by 100 epochs of fine-tuning. After each phase, a subset of the current dataset is randomly combined with the newly generated reasoning steps to form an updated synthetic dataset. The maximum number of samples selected for each problem length is detailed in Table~\ref{tab:selfgeneratedata}. 

\begin{table}[htbp]
  \centering
  \caption{Number of Samples Generated for Different Lengths}
  \vskip 0.15in
    \begin{tabular}{ccccc}
    \toprule
    \multicolumn{2}{c}{Properties} & ARI & ED & LIS \\
    \midrule
    \multirow{2}*{Phase I} & Length & 16 & 31 & 101\\
    ~ & Data Generated & 15000 & 15000 & 15000 \\
    \midrule
    \multirow{2}*{Phase II} & Length & 17 & 32 & 102\\
    ~ & Data Generated & 10000 & 10000 & 10000 \\
    \midrule
    \multirow{2}*{Phase III} & Length & 18 & 33 & 103\\
    ~ & Data Generated & 7500 & 7500 & 7500 \\
    \midrule
    \multirow{2}*{Phase IV} & Length & 19 & 34 & 104\\
    ~ & Data Generated & 6000 & 3000 & 3000 \\
    \midrule
    \multirow{2}*{Phase V} & Length & 20 & 35 & 105\\
    ~ & Data Generated & 5000 & 3000 & 3000 \\
    \bottomrule
    \end{tabular}%
  \label{tab:selfgeneratedata}%
\end{table}%


\end{document}